\pdfoutput=1

\documentclass[11pt]{article}

\usepackage[]{acl}

\usepackage{times}
\usepackage{latexsym}

\usepackage[T1]{fontenc}

\usepackage[utf8]{inputenc}

\usepackage{microtype}

\usepackage{inconsolata}
\usepackage{enumitem}

\usepackage{graphicx}
\graphicspath{ {./images/} }
\usepackage{tabularray}
\usepackage{amsmath}
 \usepackage[labelfont=bf]{caption}
\usepackage{setspace}

\title{Unveiling the Invisible: Captioning Videos with Metaphors}

\author{Abisek Rajakumar Kalarani $^\dagger$, Pushpak Bhattacharyya $^\dagger$, Sumit Shekhar $^\ddagger$\\
        $^\dagger$ Department of Computer Science and Engineering, IIT Bombay, India\\
        $^\ddagger$ Adobe Systems, India\\
        $^\dagger$ \texttt{\{abisekrk, pb\}@cse.iitb.ac.in}\\
        $^\ddagger$ \texttt{sushekha@adobe.com}}

\begin{document}
\maketitle
\begin{abstract}
Metaphors are a common communication tool used in our day-to-day life. The detection and generation of metaphors in textual form have been studied extensively but metaphors in other forms have been under-explored. Recent studies have shown that Vision-Language (VL) models cannot understand visual metaphors in memes and adverts. As of now, no probing studies have been done that involve complex language phenomena like metaphors with videos. Hence, we introduce a new VL task of describing the metaphors present in the videos in our work. To facilitate this novel task, we construct and release a manually created dataset with $\mathbf{705}$ videos and $\mathbf{2115}$ human-written captions, along with a new metric called Average Concept Distance (ACD), to automatically evaluate the creativity of the metaphors generated. We also propose a novel low-resource video metaphor captioning system: GIT-LLaVA, which obtains comparable performance to SoTA video language models on the proposed task. We perform a comprehensive analysis of existing video language models on this task and publish our dataset, models, and benchmark results to enable further research.\footnote{Code, Data, and Models are available at \url{https://github.com/abisekrk/video-metaphor-captioning}}

\end{abstract}

    

\section{Introduction}
\begin{figure}
    \centering
    \includegraphics[height=6cm, width=\linewidth]{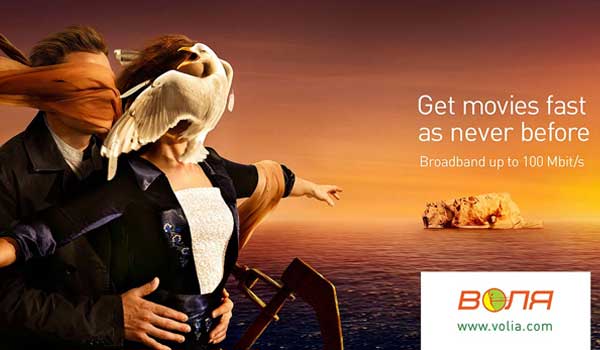}
    
    \caption{An example of a creative advertisement that shows the speed of the broadband by depicting a scene from the iconic movie `Titanic'.}
    \label{fig: image example}
\end{figure}

\begin{figure*}
    \centering
    \includegraphics[height=8cm, keepaspectratio]{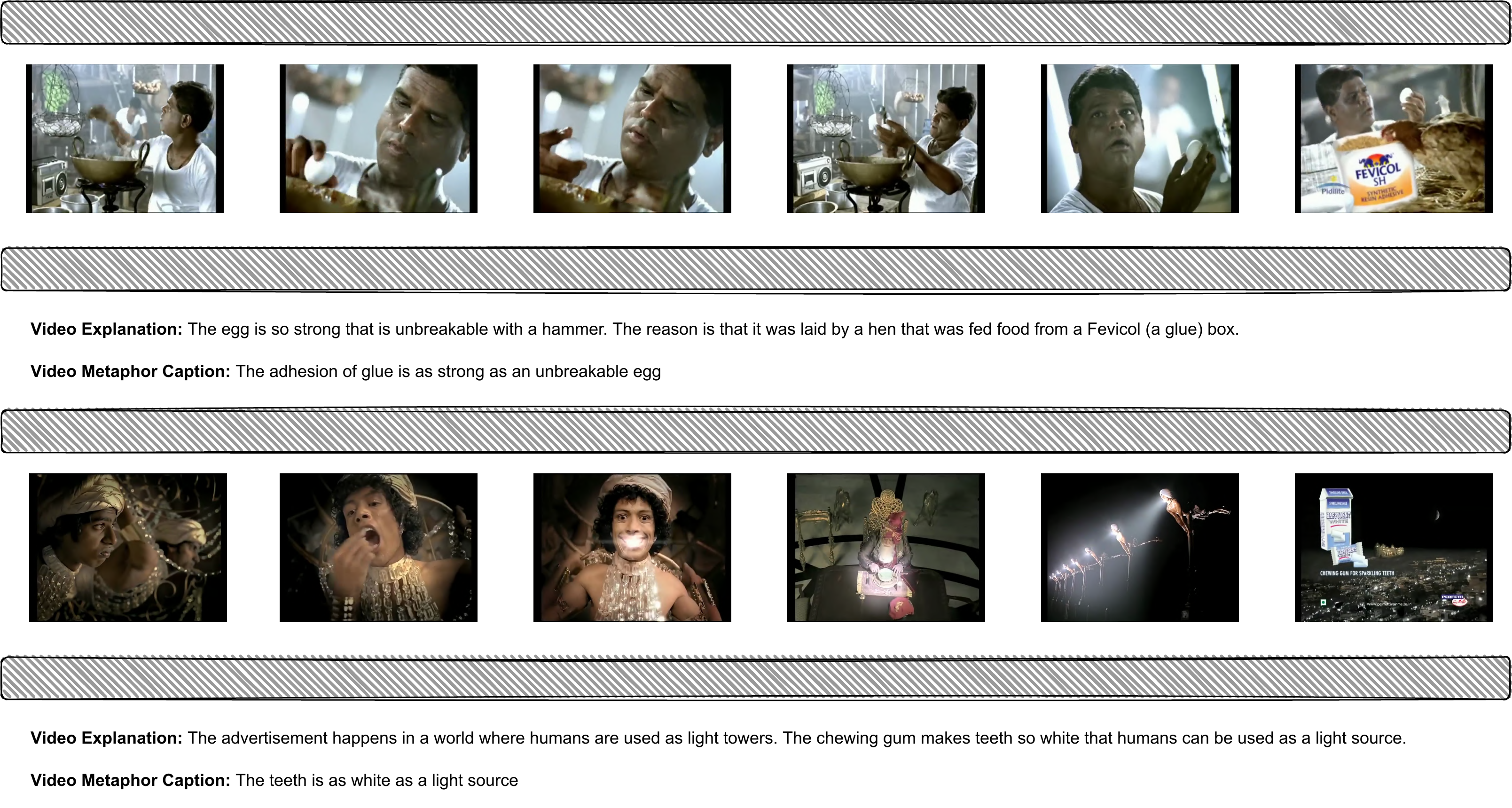}
    
    \caption{Examples of metaphors used in videos to convey ideas creatively along with their explanation}
    \label{fig: video example}
\end{figure*}

Metaphors are the most commonly used form of figurative language in literature \cite{KREUZ1993151}. Metaphors are a tool to colour the imagination of the reader by introducing unknown concepts in comparison to familiar concepts, thereby allowing them to be understood easily and powerfully. This trope is used in various creative fields like advertisements \cite{Hussain2017AutomaticUO} to convey information more effectively and includes modalities like text, images, and audio. Figure \ref{fig: image example} shows an example of using an image to creatively convey an idea. Metaphors are also used in video advertisements. Figure \ref{fig: video example} shows a few examples of how metaphors are used in video advertisements to bring emphasis to the product being advertised.

\noindent \textbf{Motivation}: Figurative languages in textual form have been well-studied in literature \cite{10.1145/3375547}. With the advent of powerful AI assistants like ChatGPT and BARD and tools that are built on top of them, it is possible to interact with these AI systems through images and audio. Hence it becomes important to build models that can handle complex language phenomena, such as metaphors, across multiple modalities. Recent works on Visual metaphors (\citealt{Yosef2023IRFLIR}, \citealt{Chakrabarty2023ISA}) focus on understanding metaphors present in images and generating images from prompts with metaphors. They show that it is challenging to deal with metaphors presented visually.

    

Recently, chat assistants that can answer questions related to videos have shown good promise on standard video datasets (\citealt{Zhang2023VideoLLaMAAI}; \citealt{Li2023VideoChatCV}; \citealt{Maaz2023VideoChatGPTTD}). However, they struggle to understand videos that contain metaphors. 
To this effect, we introduce a novel task of `Video Metaphor Captioning' (VMC) that involves describing the metaphors present in the video with a single-line caption. We manually annotate a dataset of videos with metaphor information. We build and release a novel low-resource video metaphor captioning model that achives comparable performance to SOTA video language models on video metaphor captioning task despite being trained on limited pretraining data.

Our contributions are 

\begin{enumerate}[leftmargin=*,noitemsep]
    \item A novel Vision Language (VL) task- Video Metaphor Captioning, hitherto unattempted, with a manually annotated dataset of $705$ videos comprising $2115$ captions (Section \ref{section: dataset main})
    
    \item A novel low-resource Vision-Language model (GIT video model followed by Vicuna LLM) pretrained and fine-tuned for video metaphor captioning (Section  \ref{section: model}).
    
            
    
    \item Strong baselines which are the SoTA benchmarks for the task of ``Video metaphor captioning'' (Table  \ref{tab:main_results}).
    
    \item  A new metric- Average Concept Distance (ACD) for automatically evaluating the creativity of metaphors generated by the model (Section \ref{section: metric}).

    \item Experimental results and analysis that show that existing video language models lack deeper understanding of videos to understand video metaphors (Section \ref{Section: Results}).
    
\end{enumerate}

\subsection{Problem Statement}
\textbf{Input}: Video\\
\textbf{Output}: Caption describing the metaphor.

Video metaphor captioning is the task of describing the metaphor in the video. Given a video, the model generates a single-line description of the following format: `Primary concept' is as `property' as `secondary concept'. The model should hence identify the object being compared, the object it is being compared to, and the property that links both from the video and include them in the caption.



\subsection{Background} \label{section: background}
\citet{lakoff1993contemporary} describes metaphor as a mapping between a source and target domain through shared properties. For example, in the sentence `\textit{The development has hit a wall}',  hitting a wall denotes that the development has been halted. The target domain is halting and the source domain is wall and the property of wall is used to describe halting. 

Metaphors and similes can be simplified to a syntax of A is B, where A is being compared to B. We use this simple syntax inspired from \citet{Akula2022MetaCLUETC}. A is denoted as the primary concept and B is referred to as the secondary concept. For example, in the sentence ``\textit{The blanket is as white as snow}'', the primary concept is the `blanket' and it is compared to the secondary concept `snow'. The property that links them is their `colour'. Following prior work \cite{Akula2022MetaCLUETC}, we use the following template to describe the metaphors present in the videos: \textit{Primary Concept} is as \textit{property} a \textit{Secondary Concept}

\section{Related Work}


Recently, significant efforts have been made to understand metaphors to detect and generate them. Many sentence-level and token-level datasets have been released to facilitate the same (\citealt{Birke2006ACA}; \citealt{article}; \citealt{Tsvetkov2014MetaphorDW}; \citealt{Mohammad2016MetaphorAA}; \citealt{mohler-etal-2016-introducing}).

\textbf{Metaphor Detection} is the task of classifying if the given sentence/token contains a metaphor or not. In recent years, metaphor detection has been explored with the aid of large language models. \citet{Choi2021MelBERTMD} used the contextual embeddings from BERT \cite{devlin2018bert} and RoBERTa \cite{Liu2019RoBERTaAR} with a late interaction mechanism to make use of linguistic metaphor identification theories. \citet{Aghazadeh2022MetaphorsIP} probed and analyzed the metaphorical language encoded in the large language models. \citet{Su2020DeepMetAR} used both global sentence features and POS information to perform token-level metaphor detection. \citet{Badathala2023AMM} used a multitasking approach to detect hyperbole and metaphors together.

\textbf{Metaphor generation} is the task of generating metaphorical sentences given a literal sentence (\citealt{Abe2006ACM}, \citealt{Terai2010ACS}). Metaphor generation was initially modelled as a template-filling task. \citet{veale-2016-round} used templates to generate metaphoric tweets. \citet{Stowe2020MetaphoricPG} used masked language modelling by masking the verbs in the literal sentence and training the model to replace it with its metaphoric counterparts. \citet{stowe-etal-2021-metaphor} used FrameNet embeddings to generate metaphoric sentences by replacing verbs with metaphoric verbs in literal sentences.

\textbf{Visual Metaphors}: The detection and generation of metaphors in textual form have been explored extensively but the use of metaphors in other modalities like images is not explored until very recently. \citet{Akula2022MetaCLUETC} introduced a set of tasks related to understanding visual metaphors. They showed that existing Vision-Language models are not good at understanding visual metaphors. \citet{Yosef2023IRFLIR} introduced a multimodal dataset that contains metaphors, similes, and idioms with corresponding images for them. \citealt{Zhang2021MultiMETAM}, \citealt{Hwang2023MemeCapAD}, and \citealt{Xu2022METMemeAM} explored the uses of metaphors in memes and released datasets for understanding metaphors in memes. \citet{Chakrabarty2023ISA} explored generating visual metaphors from metaphorical input sentences. They release a dataset called HAIVMet which contains 6476 images of visual metaphors generated with  DALL-E 2 \cite{Ramesh2022HierarchicalTI}. \citet{Achlioptas2021ArtEmisAL} and  \citet{Achlioptas2022AffectionLA} explore emotions invoked by images.

\textbf{Video Captioning}: Video captioning is the task of generating a single-line natural language description of the video. Video-Text models are trained on large-scale paired video and language datasets to align frames to text in the captions. \citet{Sun2019VideoBERTAJ} built on BERT \cite{Devlin2019BERTPO} model by learning a joint representation for visual and text tokens for video-text tasks. \citet{Lei2021LessIM} proposed CLIPBERT that uses sparse sampling to sample short clips from videos to learn visual representation instead of using the whole video and showed remarkable performance. \citet{Luo2020UniViLMAU} is a Unified Video and Language pre-training model for both multimodal understanding and generation built by pretraining the model on 5 diverse objectives. \citet{Zellers2021MERLOTMN} uses spatial and temporal objectives during pretraining on a large-scale dataset of videos with transcriptions to align videos to text. The GIT model \cite{Wang2022GITAG} is trained on a large corpus of parallel image-text data. It used a single image encoder and single text decoder and modeled multiple vision-text tasks as a language modeling task. These models however cannot follow instructions which makes it difficult to adapt to newer tasks.

\textbf{Video Assistants}:  Recent success in using frozen LLMs with vision encoders for instruction fine-tuning for Image-Text tasks (\citealt{Li2023BLIP2BL}; \citealt{Liu2023VisualIT}) has inspired the use of instruction fine-tuning for videos. Video-LLaMA \cite{Zhang2023VideoLLaMAAI} use frozen visual and audio encoders and projects them to the embedding space of LLMs using Q-formers as in BLIP-2 \cite{Li2023BLIP2BL}.  \citet{Li2023VideoChatCV} use information from image, video, and ASR tools along with video embedding to align video frames to text. Video-ChatGPT \cite{Maaz2023VideoChatGPTTD} use CLIP \cite{Radford2021LearningTV} as the visual encoder and Vicuna \cite{zheng2023judging} as the LLM and train the model on 100,000 video and instruction pairs. Video-LLaVa \cite{Munasinghe2023PGVideoLLaVAPG} uses a unified representation space for both images and videos. They use LanguageBind \cite{Zhu2023LanguageBindEV} to map raw features to LLM's text feature space. They obtain SOTA results on multiple vision-language tasks.


All these models are trained on large-scale video and text data. We propose a new model GIT-LLaVA that uses a frozen video foundation model with an LLM that can be fine-tuned with a few hundred videos to perform video metaphor captioning. Also, our work focuses on visual metaphors in videos which has not been explored before.

\section{Dataset} \label{section: dataset main}

No existing datasets have metaphor details available for videos. As advertisements have metaphorical representations in them to convey additional messages to viewers, we choose the Pitt’s Ads dataset \cite{Hussain2017AutomaticUO} for constructing our dataset. The Pitt’s Ads dataset consists of advertisement images and videos on a wide range of topics. The released dataset contained URLs to $3,477$ videos out of which only $2063$ videos are currently accessible. We annotate these videos with metaphor information for our experiments. Additionally, we also query YouTube with keywords like advertisements, creative advertisements, funny advertisements, etc. using the YouTube Search tool\footnote{\url{https://pypi.org/project/youtube-search-python/}}. We filter videos that are less than $2$ minutes and add them to our Video Metaphor Captioning Dataset (VMCD) if they have metaphors in them.

\subsection{Annotation Details}
We employed three annotators to annotate data for our novel task- video metaphor captioning. The annotators were given detailed explanations about metaphors and visual metaphors with examples. They were given two tests with examples consisting of metaphoric and non-metaphoric videos and asked to classify them. The annotators were shortlisted based on their ability to identify metaphors present in the videos. In our final batch of annotators, all three annotators were in the age bracket of 24-30 years. All three annotators are proficient in English with Masters degrees. Each video is annotated by all the three annotators. More details are discussed in Appendix \ref{section: Appendix annotation details}


\begin{figure*}
    \centering
    \includegraphics[height=9cm, width=\linewidth]{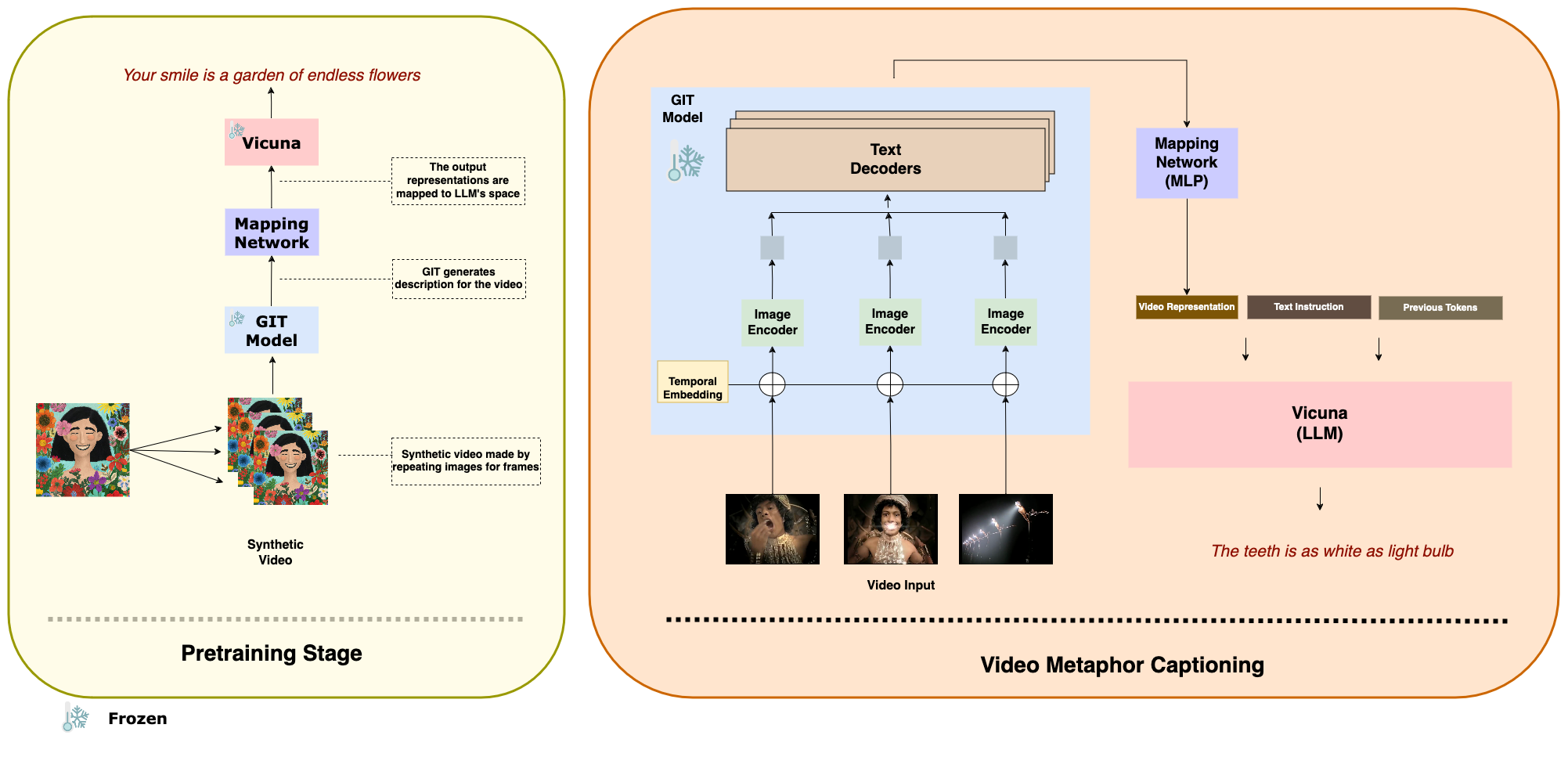}
    
    \caption{An overview of our Video Metaphor Captioning system, GIT-LLaVA. In the pretraining stage, only the mapping network is trained with image datasets. In the finetuning stage, the text decoder representation of the GIT model is mapped to the embedding space of the Vicuna model to generate metaphor captions.}
    \label{fig: git llava architecture}
\end{figure*}

\subsection{Dataset Statistics}
Interpretation of metaphors present in videos is very subjective and each annotator can understand it differently. We observed that the captions for each video were diverse. We only include videos in our final dataset that are classified as metaphors by all three annotators. This ensures that the VMC dataset has videos that are unambiguously metaphoric.

We employed an additional expert annotator who is a Masters student in English literature and proficient in understanding metaphors to validate the captions written by the three annotators. We also used the \emph{GPT-3.5-turbo} model \cite{Ouyang2022TrainingLM} to check for grammar and typos in the captions written by our annotators. The annotators were asked to rewrite the captions if any flaw was identified in terms of spelling or grammar. These quality checks ensured the quality of captions present in the dataset.

All videos are accompanied by three captions. Our \textbf{Video Metaphor Captioning Dataset (VMCD)} consists of $705$ metaphoric videos with $2115$ captions. The train, validation, and test split contain $400$, $55$, and $250$ videos each with $1200$, $165$, and $750$ captions respectively. The average duration of the video is $54$ seconds, and the average length of the caption is $8.9$ words. Figure \ref{fig: length distribution} shows histograms for the distribution of video duration and caption lengths.

\subsection{Pretraining Dataset} \label{dataset: synthetic main}
The manually annotated VMC dataset is small and is not sufficient to pretrain and finetune a model from scratch. Hence we initially pretrain our model on a larger dataset of visual metaphor images and then finetune it on the VMCD to report results. The existing image metaphor datasets- \citet{Zhang2021MultiMETAM} and \citet{Akula2022MetaCLUETC} are not publicly available and therefore cannot be used in our experiments. Hence we use the HAIVMet dataset \cite{Chakrabarty2023ISA} that contains DALLE-2 generated images for metaphor prompts as our pretraining dataset.




Our pretraining dataset contains images from the HAIVMet and MSCOCO \cite{Lin2014MicrosoftCC} datasets, consisting of metaphoric and non-metaphoric images in equal parts. For the HAIVMet dataset, the prompts to generate images are used as the caption. We use all $6476$ images and their prompts as metaphor image caption pairs from the HAIVMet dataset. We use an equivalent $6476$ image-captain pairs from the MSCOCO dataset as the non-metaphor part of the pretraining dataset.


\section{Our Model} \label{section: model}
Existing video-language models are trained on video-text parallel data that do not contain much metaphor content. Hence, in addition to analyzing their performance on our dataset, we introduce new models that are introduced to metaphors in the pretraining stage itself. In our model, the video representation is obtained through a pretrained video captioning model and prefixed with an instruction sequence to a Large Language Model (LLM). The LLM generates the caption as a sequence of tokens conditioned on the video input and the instruction. Figure \ref{fig: git llava architecture} illustrates the model architecture.

We sample `$k$' frames from the input video `$V$', where $k$ depends on the input restrictions of the video captioning model.
\begin{equation}
    V_{input} = [f^{1}, f^{2},... ,f^{k}] 
\end{equation}
where $f$ denotes each frame sampled from the video. The sampled frames are fed to the video captioning model $(C)$ whose decoder output is used as the representation for the video ($H_V$). We train a simple Multilayer Perceptron (MLP) network with parameters `$W$' to map the video representation $H_V$ to the embedding space of the LLM ($H_{R}$), similar to the LLaVA model \cite{Liu2023VisualIT}. The hypothesis is that, since the model C is already trained to decode videos to captions, a simple mapping network can learn the mapping parameters `$W$' with a smaller sample of data. We use task-specific instruction ($X_{inst}$) as input and the model is trained to reduce the cross entropy loss. The output is generated autoregressively. 

\begin{equation}
    H_{V} = C(V_{input})
\end{equation}
\begin{equation}
    H_{R} = W.H_{V}
\end{equation}
\begin{equation}
    \mathcal{L} = \sum^{n}_{i=1}logP_{\theta}(X_{i}| X_{inst}, H_{R})
\end{equation}


where $\theta$ represents the parameters of the LLM, $X_{i}$ denotes the current token being predicted. The LLM is trained with this language modeling objective. We refer to this model as `GIT-LLaVA'. We also explore a variation of GIT-LLaVA called GIT-LLaVA-X where we split the video into multiple equal-sized clips and obtain full video representation by summing up the video representation of each clip.


We use the LLaVA-13B-V1.5 \cite{Liu2023VisualIT} model architecture for our experiments inspired by its success on many Vision Language tasks. We use the Generative Image Text Transformer model (GIT) \cite{Wang2022GITAG} as the video captioning model (C) for obtaining the video representation and Vicuna \cite{zheng2023judging} as the LLM. In all our experiments we freeze the weights of the GIT model and only finetune the mapping network and the LLM. Our model has two key advantages. Since we train the mapping network to learn the mapping between the GIT decoder state to the embedding space of the LLM, the mapping network maps GIT's understanding of the video in the form of its representation to the LLM's embedding space, allowing the LLM to directly generate output from the video. This also reduces the need to pretrain the model on a huge corpus of Video-Text parallel data which is resource intensive.

\section{Experiments}
Our experiments follow a two-step process. The models are first pretrained on the pretraining data built from MSCOCO and HAIVMet datasets and then finetuned on the VMC dataset. We discuss the experiment settings for both as follows.

\subsection{Pretraining}
Our video metaphor captioning system uses a pretrained video captioning model to obtain video representation. The video representation needs to be mapped to the embedding space of the LLM for it to generate fluent captions. Our VMC dataset is small and may not be sufficient to learn this mapping. Hence, we initially pretrain the model on a pretraining dataset of metaphor and normal images, since no existing dataset has videos with metaphor information.

The images from the pretraining dataset are converted to video by repeating the images to form frames of the video. As the video model is frozen, it does not affect the video understanding abilities of our system. This synthetic video is then fed as input to the video captioning model from which the video representations are obtained. The mapping network trained on this dataset is used in the finetuning stage where video data is used.

We use the Generative Image-to-Text (GIT) model \cite{Wang2022GITAG} as our video captioning model for obtaining video representation. We use the GIT-large model that is fine-tuned for video captioning on the VaTeX dataset \cite{Wang2019VaTeXAL}. We use the Vicuna-13B model \cite{zheng2023judging} as our LLM. We pretrain the model by creating synthetic videos consisting of $6$ frames of the same image with a batch size of $4$. We pretrain the model for 2 epochs. In the pretraining stage, both GIT and Vicuna models are frozen and only the parameters of the mapping network are updated.

\subsection{Video Metaphor Captioning} \label{section: Experiment VMC}
The model is fine-tuned for video metaphor captioning on the VMC dataset after pretraining. The model is fine-tuned for $5$ epochs with early stopping on the validation set. We explore two frame selection strategies for our models. The GIT-Large model only supports video captioning with $6$ frames as input. We sample $2$ frames in temporal order across the three different parts of the video- start, middle, and end. This ensures that the $6$ frames cover the entire span of the video.

We also perform additional experiments where $6$ frames are sampled from multiple parts of the video, which we call GIT-LLaVA-X. The video is split into $4$ video clips with equal duration and video representation is obtained for each video clip using the GIT model. The final representation is obtained by summing up the representations for each video clip. Table \ref{tab:ablation results} compares the performances of models with different numbers of video clip splits.

We use a batch size of $4$ with an initial learning rate of $2e-6$ with a warmup ratio of $0.03$. Cosine Annealing is used as the learning rate scheduler. BFloat16 precision is used while training the model on 4 A100 GPUs. 

\textbf{Baselines:} We use the GIT \cite{Wang2022GITAG}, Valley \cite{Luo2023ValleyVA}, Video-ChatGPT \cite{Maaz2023VideoChatGPTTD}, and Video-LLaVa \cite{Munasinghe2023PGVideoLLaVAPG} as baselines in our experiments. GIT is chosen as the baseline as it is used as our video encoder. Video-ChatGPT and Valley have shown promising performance in following instructions in the video setting. Video-LLaVA has achieved SOTA performance on many Video-Language tasks. They also have diverse vision and language backbones and thus would make for a fair comparison. More details on baselines are discussed in Appendix \ref{section: appendix baselines}

\begin{figure*}
    \centering
    \includegraphics[width=\linewidth]{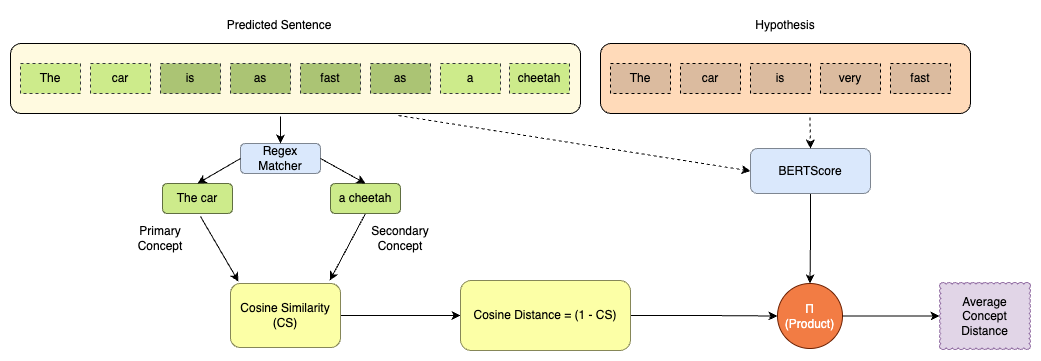}
    
    \caption{An overview of the ``Average Concept Distance'' metric. We compute the cosine distance between the primary and the secondary concept and ground it with the BERTScore.}
    \label{fig: acd metric example}
\end{figure*}

\section{Evaluation Metrics} \label{section: metric}
We evaluate the performance of our model using a set of automated metrics and human evaluation. The n-gram overlap-based metrics- BLEU \cite{papineni-etal-2002-bleu}, ROUGE \cite{lin-2004-rouge}, and CIDEr \cite{Vedantam2014CIDErCI} are commonly used to compare the performance of the model in captioning tasks. As discussed in previous works on metaphor generation, the n-gram overlap based metrics cannot capture the quality of generated metaphors. This is because the same information can be conveyed through different comparisons. Hence, we also report BERTScore \cite{Zhang2019BERTScoreET} that compares the semantic similarity of the generated caption and the reference caption.


In the task of video metaphor captioning, the model is trained to generate creative metaphors as output. Previous works rely on manual evaluation to quantify the creativity and metaphoricity of the generated captions. As no existing metric can be used to evaluate the creativity of metaphors, we introduce a new and intuitive metric called- ``Average Concept Distance'' (ACD). It is calculated as:

\setlength{\belowdisplayskip}{0pt} \setlength{\belowdisplayshortskip}{0pt}
\setlength{\abovedisplayskip}{0pt} \setlength{\abovedisplayshortskip}{0pt}



\begin{gather}
\text{CS} = \text{Cosine(PC, SC)} \\
\text{ACD} = \frac{\sum_{i}^{n}  \text{BERTScore(hyp, pred)} * (1-\text{CS}) }{n}
\end{gather}

\setlength{\belowdisplayskip}{0pt} \setlength{\belowdisplayshortskip}{0pt}
\setlength{\abovedisplayskip}{0pt} \setlength{\abovedisplayshortskip}{0pt}
where PC and SC denote the primary and secondary concepts in the predicted caption respectively and Cosine denotes the cosine similarity between them. The primary and secondary concepts denote the object of comparison and the object it is being compared to respectively.  Average Concept Distance (ACD) is obtained by weighing the cosine distance between the concepts with the BERTScore of the predicted caption. The caption `The car is as fast as a jeep' is less creative as it makes an obvious comparison while the caption `The car is as fast as a cheetah' is more creative. This can be captured by the CS metric but a disfluent caption like `The adsfd is as fast as a cdsak' will also score low on CS and this can be captured by the ACD metric. 


S-BERT \cite{reimers-2019-sentence-bert} (\emph{all-mpnet-base-v2}) is used to obtain representations for PC and SC. For captions that do not contain either PC or SC, the similarity score is set as 1 to penalize the model. Thus the model is evaluated based on how diverse comparison it can make for the object in question. Figure \ref{fig: acd metric example} provides an overview of the ACD metric computation. In addition to these automated metrics, we also manually evaluate models on four metrics- Fluency, Primary Concept Consistency, Consistency, and Creativity. 


\begin{table*}[h]
\centering
\renewcommand{\arraystretch}{1.1} 
\begin{tabular}{l|r|r|r|r|r|r} 
\hline
\textbf{Model} & \multicolumn{1}{l|}{\textbf{BLEU-4} $\uparrow$} & \multicolumn{1}{l|}{\textbf{Rouge-L} $\uparrow$} & \multicolumn{1}{l|}{\textbf{CIDEr} $\uparrow$} & \multicolumn{1}{l|}{\textbf{BERT-F1} $\uparrow$} & \multicolumn{1}{l|}{\textbf{ACS}$\downarrow$} & \multicolumn{1}{l}{\textbf{ACD}$\uparrow$}  \\ 
\hline
Video-ChatGPT & 0.38 & 3.23 & 0.03 & 0.12 & 1.00 & 0.00 \\
Valley         & 1.00                            & 14.40                               & 1.25                              & 0.50                                & 0.77                            & 0.15                             \\
GIT            & 5.85                            & 42.40                               & 7.49                              & 0.68                                & 0.40                            & 0.41                             \\

Video-LLaVA & \textbf{16.88} & 49.56 & \textbf{37.61} & 0.71 & 0.37 & 0.45 \\

\hline
GIT-LLaVA (Ours)      & 14.08                          & \textbf{50.62}                               & 24.26                             & \textbf{0.73}                                & \textbf{0.29}                            & \underline{0.52}                             \\
GIT-LLaVA-X (Ours)   & \underline{14.51}                            & \underline{50.59}                               & \underline{22.67}                             & \underline{0.74}                                & \underline{0.29}                            & \textbf{0.53}                             \\
\hline
\end{tabular}
\caption{Experimental results on our VMC dataset in comparison to other models. ACS and ACD denote the Average Concept Similarity and Average Concept Distance metric weighted by BERTScore respectively. Cosine similarity and distance is computed between the concepts compared in the metaphor caption. The best model is in bold and the next-best model is underlined. All reported scores are the mean scores of three runs.}
\label{tab:main_results}
\end{table*}

\section{Results and Analysis} \label{Section: Results}
We evaluate the models based on both automatic metrics and human evaluation.

\subsection{Automatic Metrics}

Table \ref{tab:main_results} compares the performance of our models with other baselines. Our models- GIT-LLaVA and GIT-LLava-X perform comparable to or better than other traditional video captioning models despite the smaller scale of pretraining data.  It can be seen that both GIT-LLaVA and GIT-LLava-X perform well on n-gram overlap-based metrics like BLEU-1, ROUGE-L, and CIDEr and also the BERTScore metric. This shows that it generates captions that are relatively more semantically similar to the ground truth captions than other models. 

\begin{figure}
    \centering
    \includegraphics[width=\linewidth]{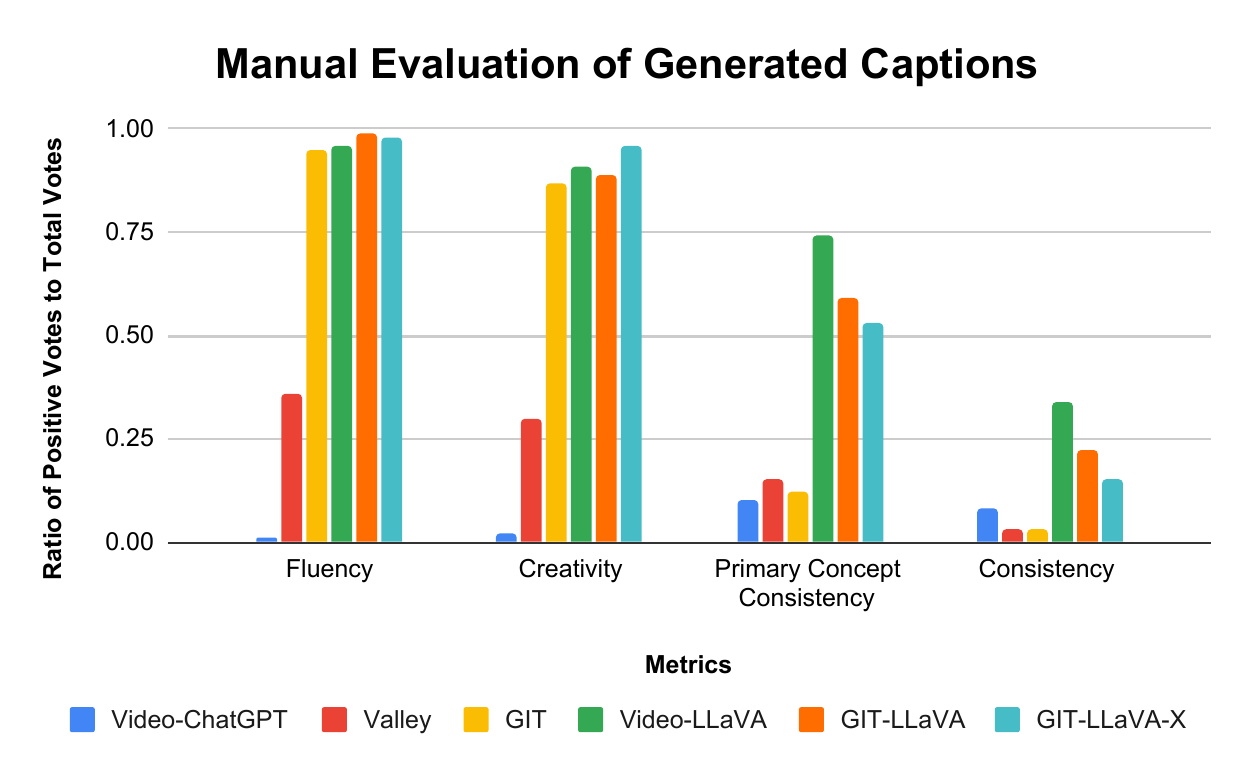}
    
    \caption{Results of human evaluation of the captions generated by our models.}
    \label{fig:human evaluation}
\end{figure}

\begin{figure*}
    \centering
    \includegraphics[height=10cm, width=\linewidth]{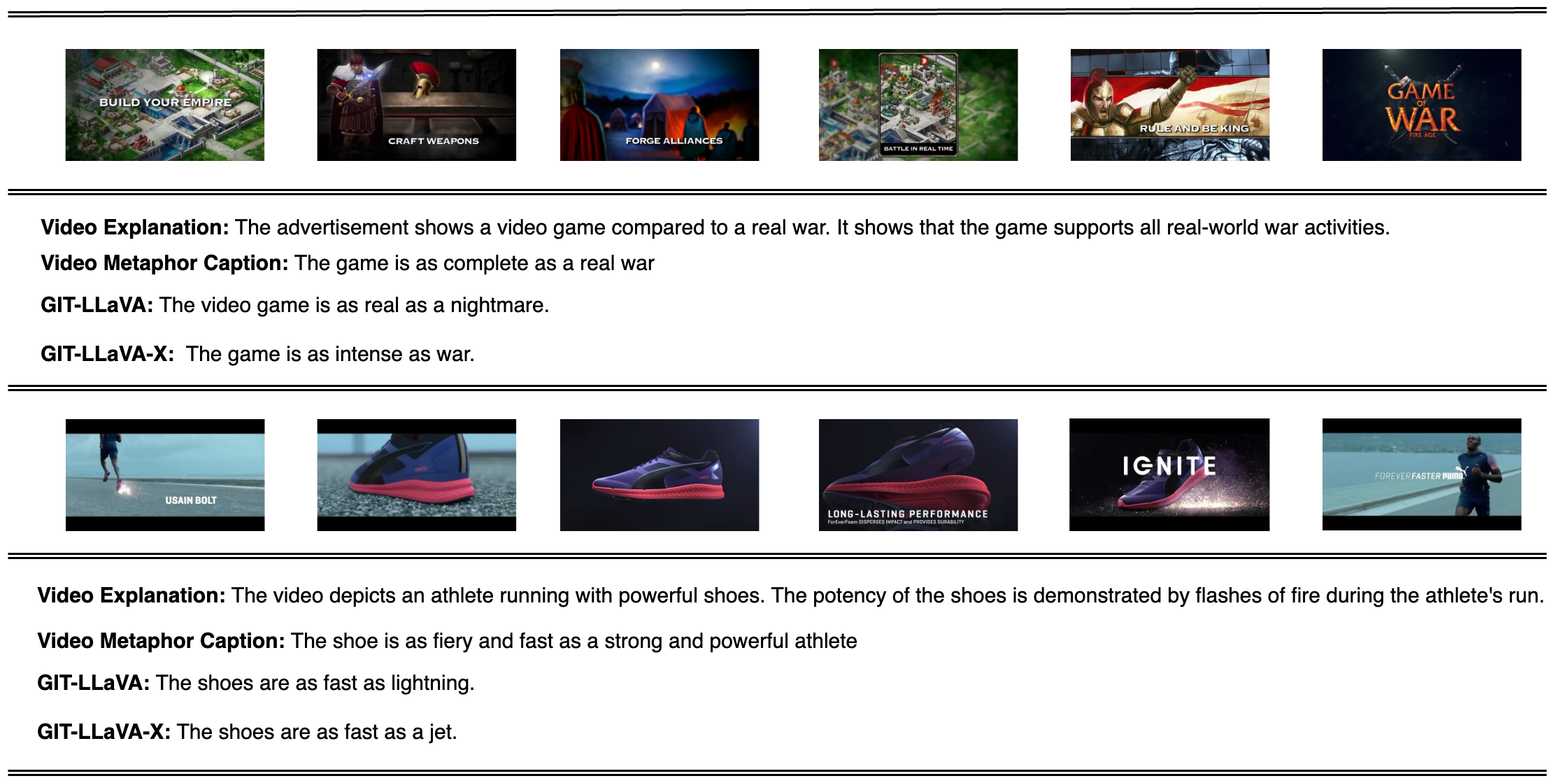}
    
    \caption{Examples of metaphor captions generated by GIT-LLaVA and GIT-LLaVA-X models.}
    \label{fig: results examples}
\end{figure*}

Figure \ref{fig: results examples} shows some examples of metaphors generated by our models. Our models achieve the best score (lowest) on the Average Concept Similarity (ACS) metric. It compares the semantic similarity of the primary and secondary concepts used in the metaphor generated. The lower scores confirm that the generated captions are creative with novel comparisons. The ACS values can also be low if the generated captions are not fluent and unrelated words are present in the caption. The Average Concept Distance (ACD) is used to capture such cases. Our models also obtain the highest scores on the ACD metric, indicating that the models generate consistent and creative captions. The best score on ACD is only $0.53$, which indicates that our system is not perfect as shown by the manual evaluation of generated captions in Figure \ref{fig:human evaluation}.

The Video-LLaVA model performs comparable to our models despite not being trained on metaphor data in the pretraining stage. It is a strong baseline as it also indirectly captures audio features from the video. Video-ChatGPT and Valley does not follow the template or generate creative captions as indicated by poor scores on all metrics. GIT generated less consistent and repeated captions. The poor performance of most baseline models is due to the low-resource nature and the inherent complexity of the task.




\subsection{Human Evaluation} \label{subsection: human evaluation}

In addition to automated metrics, we also perform human evaluation (Figure \ref{fig:human evaluation}) on 100 videos from the test set with outputs generated by all models. Three annotators in the age group of 25-30 were used to annotate these videos. Each annotator annotated 50 videos. 25 videos were common among the annotators. The manual evaluation was done on four metrics by three annotators on 100 videos. Each annotator gave binary labels for each generated caption on all 4 metrics.


\begin{itemize}[leftmargin=*,noitemsep]
\item \textbf{Fluency:} It denotes the naturalness and grammatical correctness of the generated sentence. In addition to grammatical correctness, the annotators are asked to verify if it follows the proposed template for the task.

\item \textbf{Creativity: } It denotes how creative the metaphor used in the generated caption is.

\item \textbf{Primary Concept Consistency: } It denotes if the generated caption correctly predicted the primary concept in the video. 

\item \textbf{Consistency: } It denotes if the generated caption is consistent to the full video. This checks if the primary and secondary concepts along with their relationship is captured in the caption.
\end{itemize}


\begin{table}
\centering
\small
\renewcommand{\arraystretch}{1.3} 
\begin{tabular}{l|c}
\hline
\textbf{Pairwise Comparison} & \textbf{Cohen's Kappa ($\kappa$)} \\ \hline
A vs B              & 0.639                     \\ 
B vs C             & 0.685                     \\ 
A vs C            & 0.637                     \\ \hline
\textbf{Overall Fleiss' Kappa (\textit{K})} & 0.653               \\ \hline
\end{tabular}
\caption{IAA calculations with Fleiss’ Kappa and pairwise Cohen’s Kappa among the annotators}
\label{tab:kappa}

\end{table}

The manual evaluation scores further complement the results obtained with automatic metrics. Table \ref{tab:kappa} shows the Inter-Annotator Agreement (IAA) between the three annotators for manual evaluation of captions generated by all 6 models. The scores indicate substantial agreement between the reviewers. Both sets of annotators used for annotating VMCD and the manual evaluations received fair and competitive stipends.


Our models generated mostly fluent captions but were not always consistent with the primary concept of the video. Video-LLaVA generated more consistent captions that better captured the primary concept in the video. This is primarily because of the generalizability of the model due to large pretraining data and yet it was consistent to the video less than 50\% of the time. Our GIT-LLaVA-X model was the most creative of all models. GIT generated less consistent captions. Video-ChatGPT struggled to generate anything useful. Valley generated captions that were not always following the template. Both set of evaluations indicate that our models perform comparable to Video-LLaVA despite being trained with smaller datasets.

We also compute the correlation of the Average Concept Distance (ACD) metric with the human evaluation of captions. We compared the ACD metric scores computed by our module with the binary labels provided by our annotators for the above 100 videos. The ACD scores and binary labels had a Pearson correlation coefficient of $0.403$ with p-value $<< 0.0001$. As creativity is a subjective metric, the moderate correlation is very significant.



\subsection{Error Analysis}
The most common case of error is the misprediction of the primary concept in the video as can be seen in Figure \ref{fig:human evaluation}. Figure \ref{fig: errors example} illustrates a few examples of misprediction. In the first example, the GIT-LLaVA models generate a metaphor about cars when the actual metaphor was about getting a car loan. It was also observed that videos related to shoe brands typically present more about the game and the athletes than about the shoes themselves. This leads to models generating metaphors about people and the game than about the shoes. It was also observed that videos that contain animated objects are confused for advertisements about video games resulting in metaphors being generated about video games. In general, all the models don't seem to have the ability to deeply reason about the video to generate accurate metaphors as shown by the performances on the VMC dataset.

\section{Conclusion and Future work}

In this work, we proposed a novel Vision-Language (VL) task called video metaphor captioning that probes the language reasoning abilities of the video language models. We constructed and released a manually annotated dataset for the proposed task. We also released a new metric to automatically evaluate the creativity of the generated metaphor captions. Our low-resource VL model that used a frozen video captioning model (GIT) with an LLM decoder (Vicuna) to generate metaphor captions showed comparable performance to SOTA video language models on the video metaphor captioning task. It was observed that all the video language models studied in the work lack a deeper understanding of video and language for a complex task like video metaphor captioning. We believe that our work will enable future research in this direction with our dataset and models being a strong benchmark for progress.




\section{Limitations}


We briefly describe the identified limitations in our work.

\begin{itemize}[leftmargin=*,noitemsep]
    \item \textbf{No Audio Support:} The scope of our work is only limited to understanding visual metaphors in videos. The models introduced in our work- GIT-LLaVA and GIT-LLaVA-X do not have support for audio and cannot understand metaphors introduced through audio. The audio signals like music and dialogues can be used to better understand metaphor information in videos and we intend to do this in the future.

    \item \textbf{Template Captions:} The captions in our VMC dataset follow a fixed template inspired from MetaCLUE dataset \cite{Akula2022MetaCLUETC}. This is consistent with earlier works on textual metaphor generation (\citealt{Abe2006ACM}; \citealt{Terai2010ACS}).

    \item \textbf{ACD metric for general captions: } The ACD metric involves identifying primary and secondary concepts in the caption to score the novelty of comparison. In our work, it is easy to identify the concepts due to the nature of the template. In free-form text generation tasks, it will involve an additional step of identifying primary and secondary concepts from the text. This can be done by training LLMs to identify primary and secondary concepts from the input but is beyond the scope of this work.
\end{itemize}

\section{Ethical Considerations}
We build our Video Metaphor Captioning (VMC) dataset based on the Pitt’s Ads dataset. The original dataset has links to YouTube videos and may contain some videos that propagate biases seen in advertisements. We ensure that no personal information is included in the captions written by our annotators. We also ensure that brand names are replaced with common nouns such that no identifiable information is present in our dataset. Our model uses Vicuna as the decoder and may propagate the biases held by the LLM. We urge the research community to use our models and datasets with necessary caution in downstream tasks for the same reason and use them responsibly.

\bibliography{anthology,custom}

\appendix

\section{Appendix} 
\subsection{Annotation Details} \label{section: Appendix annotation details}
The VMC dataset consists of three manually written captions for each video. The annotators were asked the following questions for each video:

\begin{enumerate}[label=\alph*), leftmargin=*,noitemsep]
    \item Does this video contain a visual metaphor?
    \item Is audio of the video required to understand the metaphor?
    \item What part of the video contains the metaphor?
    \item What is the primary concept in this video?
    \item What is the secondary concept in this video?
    \item What is the common property of both concepts?
    \item Give a one-line description of the form ``\emph{primary\_concept}'' is as ``\emph{property}'' as ``\emph{secondary\_concept}''. 
    \item A free-form description of the video.
\end{enumerate}

Questions a and b are Yes/No questions. The annotators write the time of occurrence of the metaphor in the video for question c. Question g follows the format used for annotation in the MetaCLUE dataset \cite{Akula2022MetaCLUETC} for visual metaphor in images. The VMC dataset consists of videos that were marked as metaphors (Quesetion: a) by all three annotators. We instruct annotators to ensure that no identification information is included in the primary and secondary concepts and to use common words in their place. For example, instead of `The coke is as cool as Messi in the finals', the caption is written as `The drink is as cool as the football player in the finals'.

\begin{table*}
\centering
\renewcommand{\arraystretch}{1.1} 

\begin{tabular}{l|c|r|r|r|r|r|r} 
\hline
\textbf{Model}        & \multicolumn{1}{l|}{\textbf{\# of VC}} & \multicolumn{1}{l|}{\textbf{BLEU-4}} & \multicolumn{1}{l|}{\textbf{Rouge-L}} & \multicolumn{1}{l|}{\textbf{CIDEr}} & \multicolumn{1}{l|}{\textbf{BERT-F1}} & \multicolumn{1}{l|}{\textbf{ACS}} & \multicolumn{1}{l}{\textbf{ACD}}  \\ 
\hline

1) LLaVA 7B & 1 & 10.27 & 46.21 & 18.93 & 0.69 & 0.39 & 0.43 \\
2) LLaVA 13B & 1 & 12.18 & 47.88 & 22.44 & 0.70 & 0.42 & 0.41 \\

\hline

3) GIT-LLaVA-Syn & 1 & 12.39 & 49.92 & 21.65 & 0.73 & 0.32 & 0.49 \\

4) GIT-LLaVA-X-Syn & 1 & 9.32 & 48.31 & 11.45 & 0.71 & 0.35 & 0.46 \\

5) GIT-LLaVA-NP & 1                                            & 0.72                            & 20.21                               & 1.87                              & 0.42                                & 0.99                            & 0.00                           
\\
\hline

6) GIT-LLaVA-X           & 2                                            & 11.22                           & 49.21                               & 16.60                             & 0.72                                & 0.32                            & 0.48                             \\ 

7) GIT-LLaVA-X           & 4                                            & 9.32                            & 48.31                               & 11.45                             & 0.71                                & 0.35                            & 0.46                             \\ 

8) GIT-LLaVA-X           & 6                                            & 7.29                            & 47.74                               & 8.25                              & 0.70                                & 0.33                            & 0.47                             \\ 
\hline

\end{tabular}

\caption{Ablation study results with different number of video clip segmentations. \# of VC denotes the number of video clips. GIT-LLaVA-NP denotes the model that was not pretrained on synthetic data}
\label{tab:ablation results}

\end{table*}

\begin{figure*}
    \centering
    \includegraphics[height=5cm,width=\linewidth]{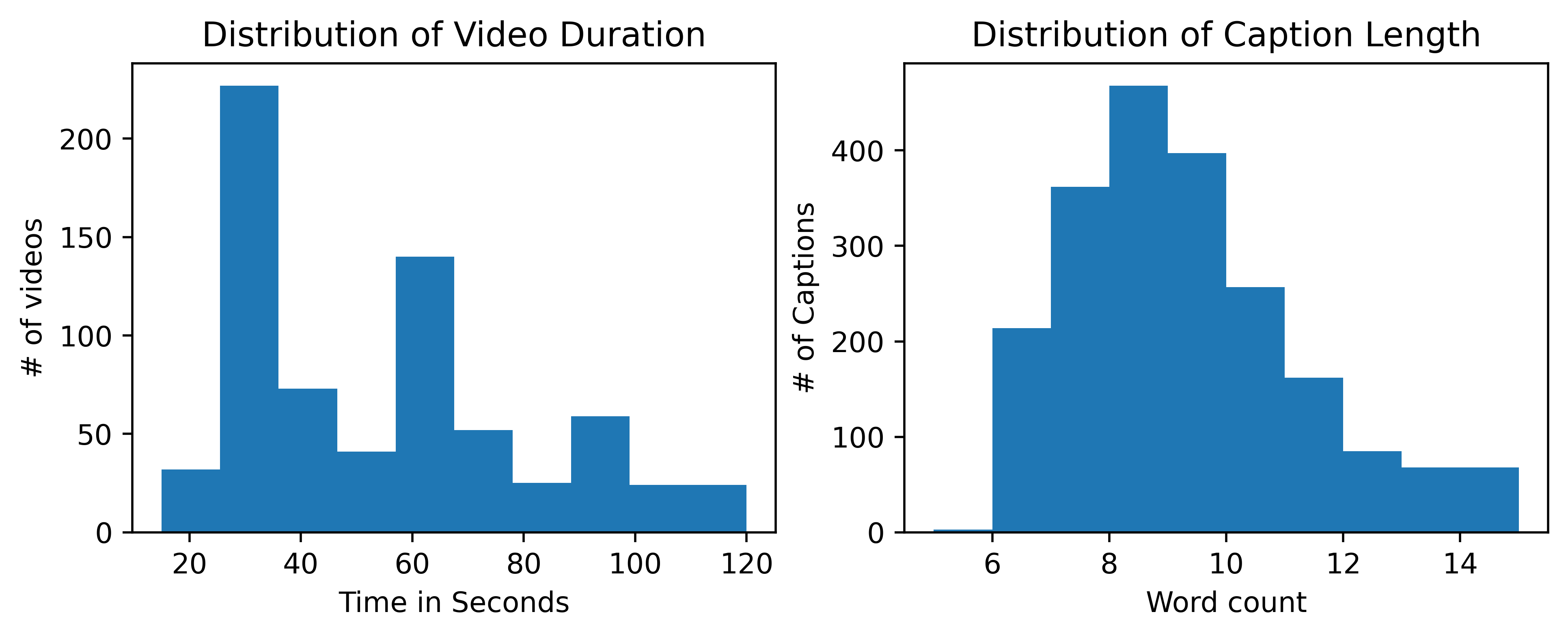}
    
    \caption{The distribution of video clip duration and caption length in the VMC dataset}
    \label{fig: length distribution}
\end{figure*}

\subsection{Baselines} \label{section: appendix baselines}
We use the GIT \cite{Wang2022GITAG}, Valley \cite{Luo2023ValleyVA}, Video-ChatGPT \cite{Maaz2023VideoChatGPTTD}, and Video-LLaVa \cite{Munasinghe2023PGVideoLLaVAPG} as our baseline models.

\textbf{GIT}: We finetune the GIT model that is already fine-tuned for video captioning on VaTEx dataset on our VMC dataset. The model is fine-tuned with a batch size of $4$.

\textbf{Video-ChatGPT}: We use the 7B model of Video-ChatGPT that is trained on $100,000$ videos. We finetune this model on our VMC dataset. The spatio-temporal features of the video are precomputed with CLIP and used during training. We use a batch size of $4$ and train it for 50 epochs with learning rate $2e-5$.

\textbf{Valley}: Valley is a video-assistant build on top of the LLaVA model. We use the Valley-2 7B model that is finetuned on video instruction data. We finetune this model on the VMC dataset with $4$ as the batch size.

\textbf{Video-LLaVA}: We use the 7B model of Video-LLaVA that is trained on image and video data. We use a batch size of $4$ and train it for 20 epochs with a learning rate $2e-5$ with default settings for other parameters.

\subsection{Prompts for Training}
As discussed in Section \ref{section: model}, the input to LLM consists of of prompt and video representation. The synthetic videos generated from MSCOCO dataset are accompanied by the prompt `\textit{What caption can best describe the video?}'. In all our experiments, the synthetic videos generated from the HAIVMet dataset and the videos from VMCD are accompanied by the same prompt, `\textit{What caption can best describe the metaphor in the video?}', during both the pretraining and finetuning stages.


\subsection{Ablation Study}
We perform different ablation studies to test the difficulty of dataset and the alternate architecture choices for our models.

\subsubsection{Image Models}

We perform experiments with LLaVA model \cite{Liu2023VisualIT} on the VMC dataset. The LLaVA 7B and 13B models are finetuned with a randomly sampled image as input to the model. The scores are reported in rows 1 and 2 of Table \ref{tab:ablation results}. The scores indicate that the metaphor present in the video cannot be understood by looking at only a single frame. This shows that the dataset is challenging and the captions makes use of the entire video.

\subsubsection{Synthetic Dataset}

Our pretraining dataset is composed of images from MSCOCO and HAIVMet. We also explored if a larger sample of synthetically generated data will help in pretraining the model better.

We use images and captions from the MSCOCO dataset \cite{Lin2014MicrosoftCC}. We prompt \emph{GPT-3.5-turbo} model with the following prompt: ``Convert the following image caption to a metaphoric image caption in the following format <primary concept> is as <property> as <secondary concept>. Input: {mscoco\_caption}''. For example, we convert the image caption `\textit{A bicycle replica with a clock as the front wheel}' to `\textit{A timepiece is as cyclical as a bicycle's revolution}'. It is important to note that the MSCOCO dataset does not contain metaphoric images in them. We generate metaphoric captions for normal images, as the goal is only to learn the mapping between literal captions and metaphoric captions.

The generated captions were then cleaned to remove captions that did not follow the template in the prompt. The final pretraining dataset consists of $90886$ images and corresponding synthetically generated metaphoric captions.  We further evaluate the quality of the generation by manual evaluation. We employed two annotators to annotate the fluency, creativity and consistency of a randomly sampled 500 generated captions. The annotators provided binary classification labels. The captions were $98.7$\% fluent, $97.8$\% creative, and $96.4\%$ consistent. This confirms the quality of the synthetic data generated by the model.

Rows 3 and 4 in Table \ref{tab:ablation results} show results of the GIT-LLaVA and the GIT-LLaVA-X models trained on this synthetic data. The results are comparable to the models trained on the previously discussed pretraining data. No improvement in the performance was observed. Since the pretraining data had metaphoric captions to normal images, it leads to increase in hallucinations during finetuning stage as the model generated metaphors about things not present in the videos.

\subsubsection{No Pretraining}

We study the impact of the pretraining stage by directly finetuning the GIT-LLaVA model on the VMC dataset. Rows 5 of Table \ref{tab:ablation results} reports the results of model finetuned without pretraining experiment. The poor performance shows that imparting metaphor knowledge in the pretraining stage is essential for model performance as the training data is smaller.

\subsubsection{Additional Video Components}
We perform an ablation study on the number of video clip segments that can be fed as input to the video captioning model. We split the video into 1, 2, 4, and 6 parts and fed the video clips to the GIT model. The final video representation is obtained by summing up the individual clip representations. The models were trained as discussed in Section \ref{section: Experiment VMC}. Table \ref{tab:ablation results} shows the results of the ablation study. On comparing the performance of these models with GIT-LLaVA, it can be seen that adding more video clips did not improve the model performance.



\begin{figure*}
    \centering
    \includegraphics[height=0.4\paperheight, width=\linewidth]{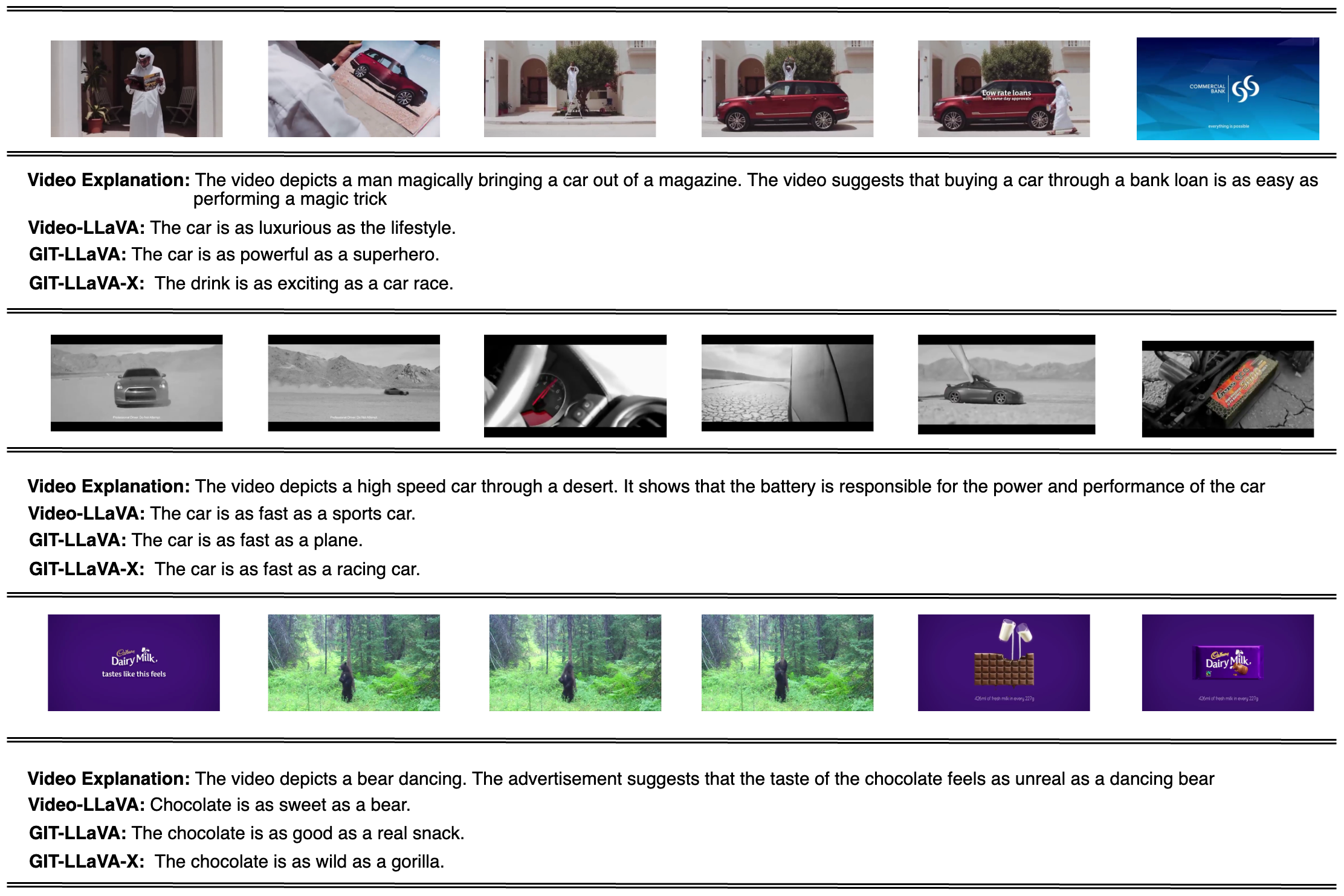}
    
    \caption{Examples of prediction mistakes done by the models on video metaphor captioning}
    \label{fig: errors example}
\end{figure*}


\label{sec:appendix}


\end{document}